\documentclass[fleqn,10pt]{wlscirep}
\usepackage{hyperref}

\title{Segmentation of the Proximal Femur from MR Images using Deep Convolutional Neural Networks}

\author[1,2,*]{Cem M. Deniz}
\author[3]{Siyuan Xiang}
\author[4]{Spencer Hallyburton}
\author[2]{Arakua Welbeck}
\author[2]{James S. Babb}
\author[5]{Stephen Honig}
\author[3,6]{Kyunghyun Cho}
\author[1]{Gregory Chang}
\affil[1]{Department of Radiology, New York University School of Medicine, New York, NY 10016 USA}
\affil[2]{Bernard and Irene Schwartz Center for Biomedical Imaging, New York University School of Medicine, NY 10016 USA}
\affil[3]{Center for Data Science, New York University, New York, NY 10012 USA}
\affil[4]{Harvard College, Cambridge, MA 02138 USA}
\affil[5]{Osteoporosis Center, Hospital for Joint Diseases, New York University Langone Medical Center, New York, NY 10003 USA}
\affil[6]{Courant Institute of Mathematical Science, New York University, New York, NY 10012 USA}

\affil[*]{cem.deniz@nyumc.org}


\begin{abstract}

This is a pre-print of an article published in Scientific Reports. The final authenticated version is available online at: \href{https://doi.org/10.1038/s41598-018-34817-6}{https://doi.org/10.1038/s41598-018-34817-6}.

Magnetic resonance imaging (MRI) has been proposed as a complimentary method to measure bone quality and assess fracture risk. However, manual segmentation of MR images of bone is time-consuming, limiting the use of MRI measurements in the clinical practice. The purpose of this paper is to present an automatic proximal femur segmentation method that is based on deep convolutional neural networks (CNNs). This study had institutional review board approval and written informed consent was obtained from all subjects. A dataset of volumetric structural MR images of the proximal femur from 86 subject were manually-segmented by an expert. We performed experiments by training two different CNN architectures with multiple number of initial feature maps and layers, and tested their segmentation performance against the gold standard of manual segmentations using four-fold cross-validation. Automatic segmentation of the proximal femur achieved a high dice similarity score of 0.94$\pm$0.05 with precision = 0.95$\pm$0.02, and recall = 0.94$\pm$0.08 using a CNN architecture based on 3D convolution exceeding the performance of 2D CNNs. The high segmentation accuracy provided by CNNs has the potential to help bring the use of structural MRI measurements of bone quality into clinical practice for management of osteoporosis. 
\end{abstract}
\begin{document}

\flushbottom
\maketitle
%
%
\thispagestyle{empty}

\section*{Introduction}
Osteoporosis is a public health problem characterized by increased fracture risk secondary to low bone mass and microarchitectural deterioration of bone tissue. Hip fractures have the most serious consequences, requiring hospitalization and major surgery in almost all cases. Early diagnosis and treatment of  osteoporosis plays an important role in preventing osteoporotic fracture. Bone mass or bone mineral content is currently assessed most commonly via dual-energy x-ray absorptiometry (DXA) \cite{Genant1996,Cummings2002}. Over the years, cross-sectional imaging methods such as quantitative computed tomography (qCT) \cite{Okazaki2014,Chiba2013,Bousson2006a,Nagarajan2015,Boutroy2005,Kazakia2007,Muller1994} and magnetic resonance imaging (MRI) \cite{Link1998,Majumdar1999,Wehrli1998,Majumdar2002,Wehrli1993} have been shown to provide useful additional clinical information beyond DXA secondary to their ability to image bone in 3-D and provide metrics of bone structure and quality \cite{Link2012}. 

MRI has been successfully performed \textit{in vivo} for structural imaging of trabecular bone architecture within the proximal femur \cite{Krug2005,Han2014,Chang2014a}. MRI provides direct detection of trabecular architecture by taking advantage of the MR signal difference between bone marrow and trabecular bone tissue itself. Osteoporosis related fracture risk assessment using MR images requires image analysis methods to extract information from trabecular bone using structural markers, such as topology and orientation of trabecular networks \cite{Hildebrand1999,Ladinsky2008,Gomberg2000}, or using finite element (FE) modeling \cite{Rajapakse2012,MacNeil2008,Cody1999}. Bone quality metrics derived from FE analysis of MR images are shown to correlate with high resolution qCT imaging, and may reveal different information about bone quality than that provided by DXA \cite{Chang2014a}. These technical developments overlay the significance of image analysis tools to determine osteoporosis related hip fracture risk.

Initial studies of MRI assessment of bone quality in proximal femur focused on quantification of parameters within specific regions of interest (ROI), such as the femoral neck, femoral head, and Ward’s triangle, for extracting fracture risk relevant parameters \cite{Chang2014a}.  More recently, investigation of the whole proximal femur has been proposed as a way to assess the mechanical properties or strength of the whole proximal femur, rather than just a subregion \cite{Orwoll2009,Chang2014c,Rajapakse2016}. The latter, however, requires manual segmentation of the whole proximal femur \cite{Chang2014a,Carballido-Gamio2013} on MR images by an expert. Given the large number of slices for a single subject acquired by MRI during a scan session, time-consuming manual segmentation of proximal femur can hinder the practical use of MRI based hip fracture risk assessment. In addition, manual segmentation may be subject to inter-rater variability. Automatic segmentation of the whole proximal femur would help overcome these challenges. 

In previous studies, hybrid image segmentation approaches including thresholding and 3D morphological operations \cite{Zoroofi2001} as well as statistical shape models \cite{Schmid2011} and deformable models \cite{Schmid2008,Arezoomand2015} have been used to segment the proximal femur from MR images. These approaches developed automated segmentation frameworks based on sophisticated algorithms. Even though these frameworks achieve sensitivities $\sim$0.88 \cite{Zoroofi2001}, their use is limited by the time required to obtain proximal femur segmentations and by the robustness on a large variation of femur shapes.

The use of convolutional neural networks (CNNs) has revolutionized image recognition, speech recognition and natural language processing \cite{LeCun2015}. Deep CNNs have recently been successfully used in medical research for image segmentation \cite{Prasoon2013,Pereira2016,Cheng2016,Lai2015,Roth2015} and computer aided diagnosis \cite{Roth2014,Wang2015,Lu2016,Yan2016}. In contrast to previous approaches of segmentation which rely on the development of hand-crafted features \cite{Zoroofi2001, Schmid2008, Arezoomand2015}, deep CNNs learn increasingly complex features from data automatically. The first applications of CNNs in medical image segmentation used pyramidal CNN architectures \cite{Ciresan2012,Zhang2015,Dou2016,Kleesiek2016,Prasoon2013,Roth2014,Roth2015,Pereira2016,Hallyburton2017a} based on the information from local regions around a voxel as an input (patches) to predict whether the central voxel of the input patch belongs to a foreground or not. In a study using structural MRIs, Hallyburton et al. used pyramidal CNN architectures for segmenting the proximal femur to achieve moderate segmentation results \cite{Hallyburton2017a}. These approaches are limited by the size of the receptive field of the networks and by the time required for CNN training and inference, especially for volumetric datasets. Developments in image segmentation using fully convolutional network architectures have emerged resulting in more accurate pixel-wise segmentations \cite{Badrinarayanan2015,Noh2015,Ronneberger2015}. These networks used encoder-decoder type architectures, where the role of the decoder network being to project the low resolution encoder feature maps to high resolution feature maps for pixel-wise classification. Encoder-decoder based CNN architectures have been recently used extensively in the biomedical field providing accurate image segmentation \cite{Cicek2016,Fedorov2016,Milletari2016,Christ2016,Kayalibay2017,Lieman-Sifry2017,Liu2018}. 

In this work, we propose to investigate different CNN architectures based on the U-net \cite{Ronneberger2015} and  the 3D extension of the U-net, and compare their performance for automatic segmentation of the proximal femur on MR images against the reference standard of expert manual segmentation. 

\section*{Results}
\subsection*{Comparison of CNN Performance}
\begin{figure}[ht]
\centering
\includegraphics[width=\linewidth]{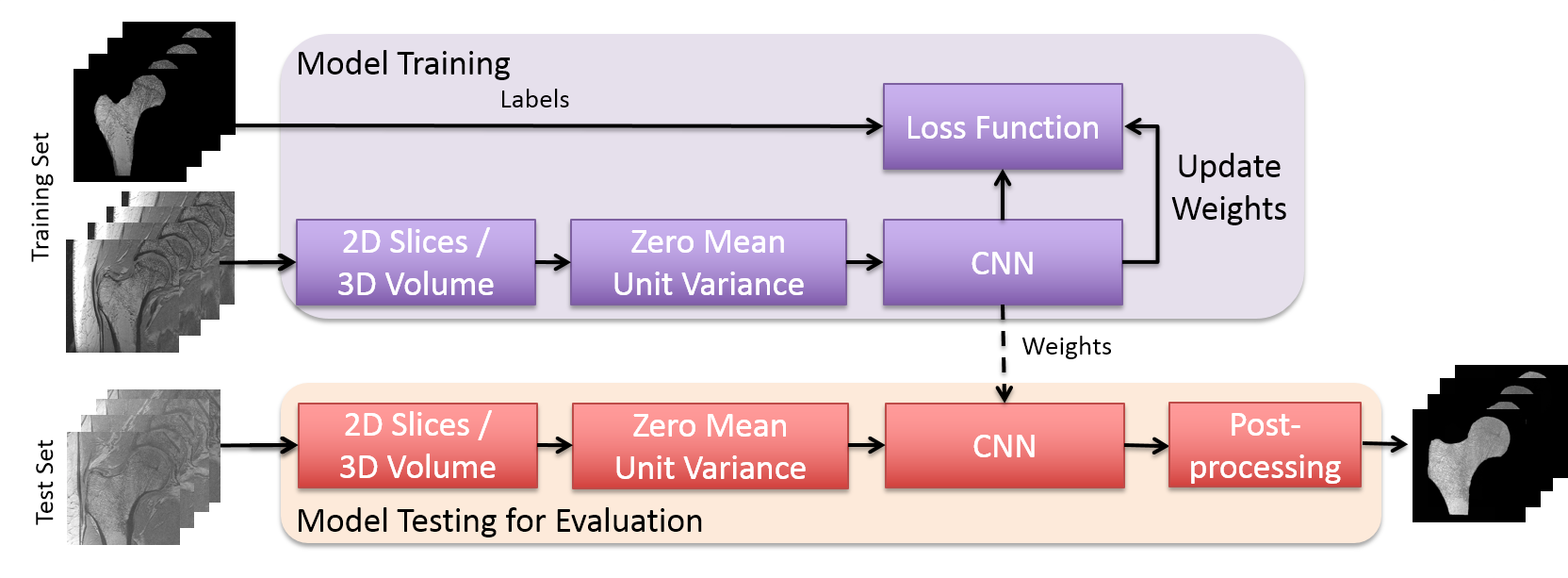}
\caption{Overview of the proposed learning algorithm for an automatic segmentation of the proximal femur. Training CNN yields automatic proximal segmentation model that is used in model evaluation on a test dataset. The output of the model is the probability of the bone which is used to obtain the proximal femur segmentation mask using a threshold.}
\label{figure:1}
\end{figure}

\begin{figure}[ht]
\centering
\includegraphics[width=\linewidth]{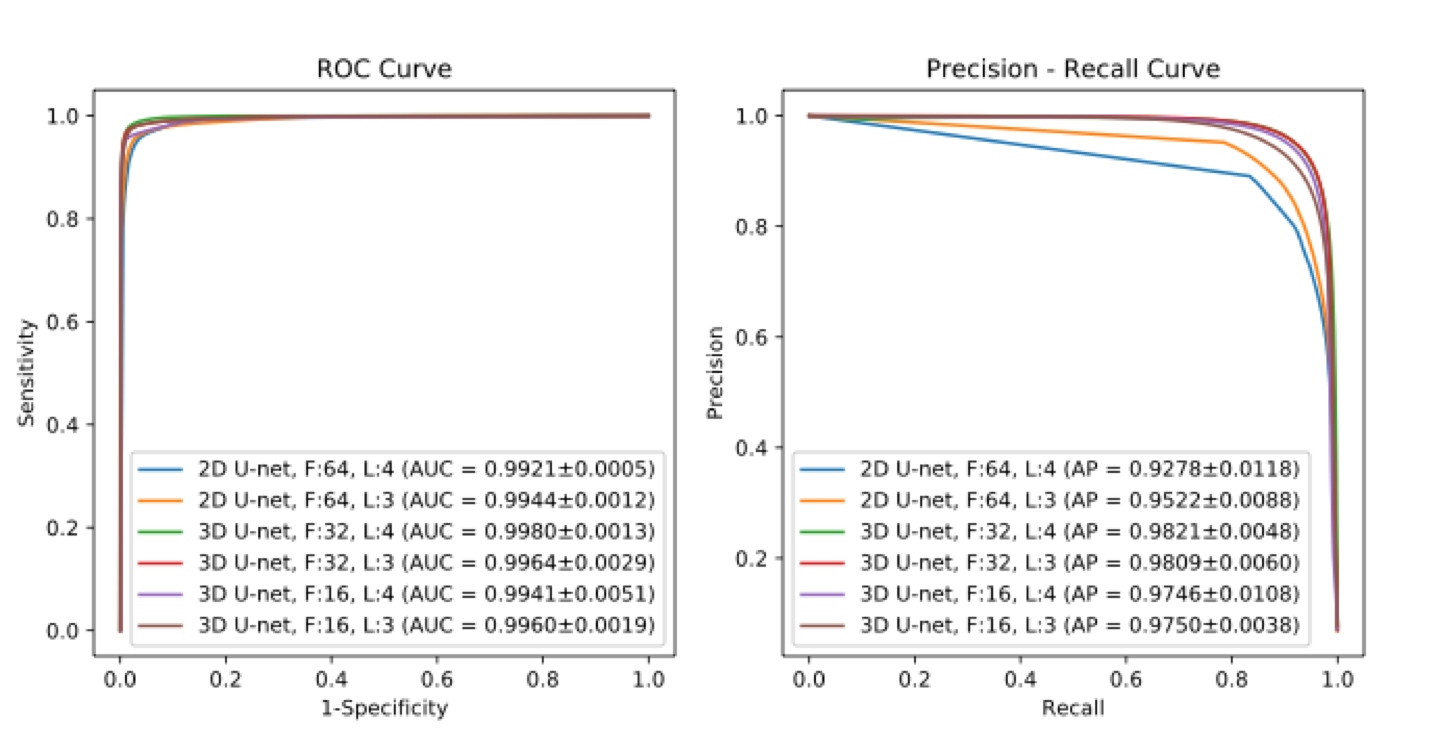}
\caption{ROC and Precision-Recall Curve for 2D and 3D CNN segmentation models. Left panel shows the receiver operating characteristics (ROC) curves of different CNNs modeled in this work. The number of initial feature maps (F) and layers (L) in the contracting/expanding paths are presented in the legend with the area under the curve (AUC). Right panel shows the precision- recall curves of modeled CNNs. In the legend, average precision (AP) is presented for comparison of different models.}
\label{figure:3}
\end{figure}

Various CNN architectures have been used for automatic segmentation of biomedical images \cite{Prasoon2013,Pereira2016,Cheng2016,Lai2015,Roth2015}. In this study, two different supervised deep CNN architectures based on 2D convolution (2D CNN) and 3D convolution (3D CNN) were used and evaluated for automatic proximal femur segmentation on MR images. An overview of the proposed approach for automatic segmentation of the proximal femur is presented in Figure \ref{figure:1}. Receiver operating characteristics (ROC) and precision-recall curve (PRC) analysis of modeled CNNs on the dataset are presented in Figure \ref{figure:3} using the mean curves from 4-fold cross-validation. The 3D CNN with 32 initial feature maps and 4 layers each in the contracting/expanding paths outperformed the other CNNs with area under the ROC curve (AUC) = 0.998$\pm$0.001 and area under the PRC curve (AP: average precision) = 0.982$\pm$0.005. This model achieved the highest accuracy on the segmentation of the proximal femur and it exceeded the performance of 2D CNNs which achieved AUC = 0.994$\pm$0.001 and AP =0.952$\pm$0.001.

\begin{figure}[ht]
\centering
\includegraphics[width=\linewidth]{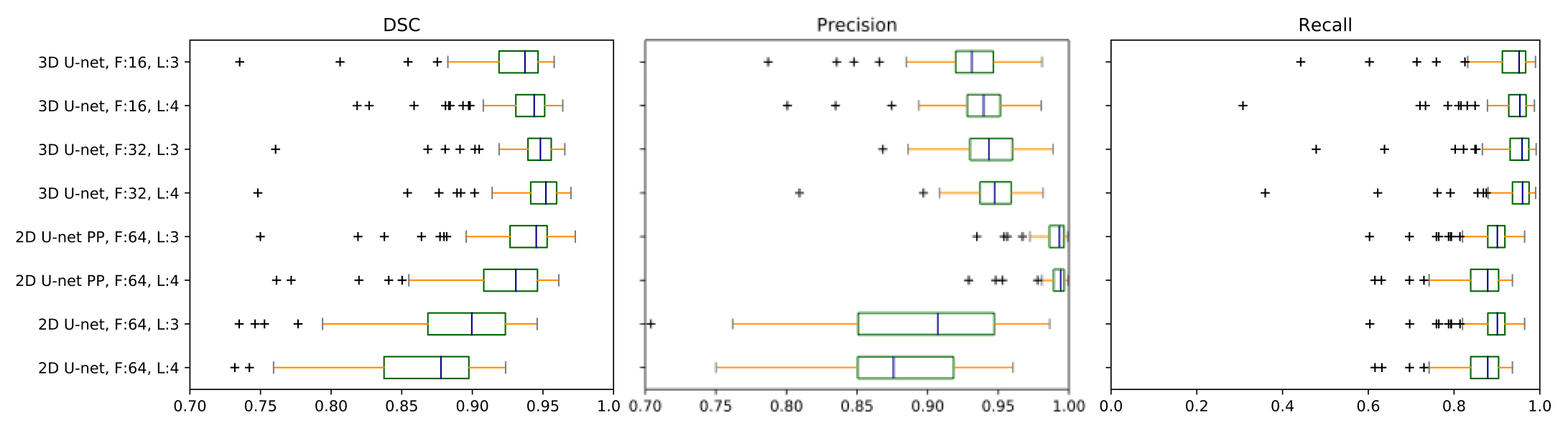}
\caption{Box plots for dice score, precision and recall. F is the number of initial feature maps, L is the number of layers, PP is the post-processing.}
\label{figure:4}
\end{figure}

The optimal threshold was applied to the segmentation probability maps to calculate a binary segmentation mask. In the 2D CNN, post-processing was also applied to the segmentation mask. The 3D CNN with 32 initial feature maps and 4 layers resulted in the highest DSC = 0.940$\pm$0.054 with precision = 0.946$\pm$0.024, and recall = 0.939$\pm$0.081. Analysis of performance metrics on individual subjects is illustrated in Figure \ref{figure:4} and Table \ref{table:2}. Applying post-processing on the 2D CNN segmentation results improved the overall accuracy of the segmentation masks as indicated by the increase in DSC on average by 7\%. As indicated by Figure \ref{figure:4}, post-processing improves the precision on 2D CNNs; however, average recall was not affected by the post-processing significantly. 

\subsection*{Segmentation Accuracy}
\begin{figure}[ht]
\centering
\includegraphics[width=\linewidth]{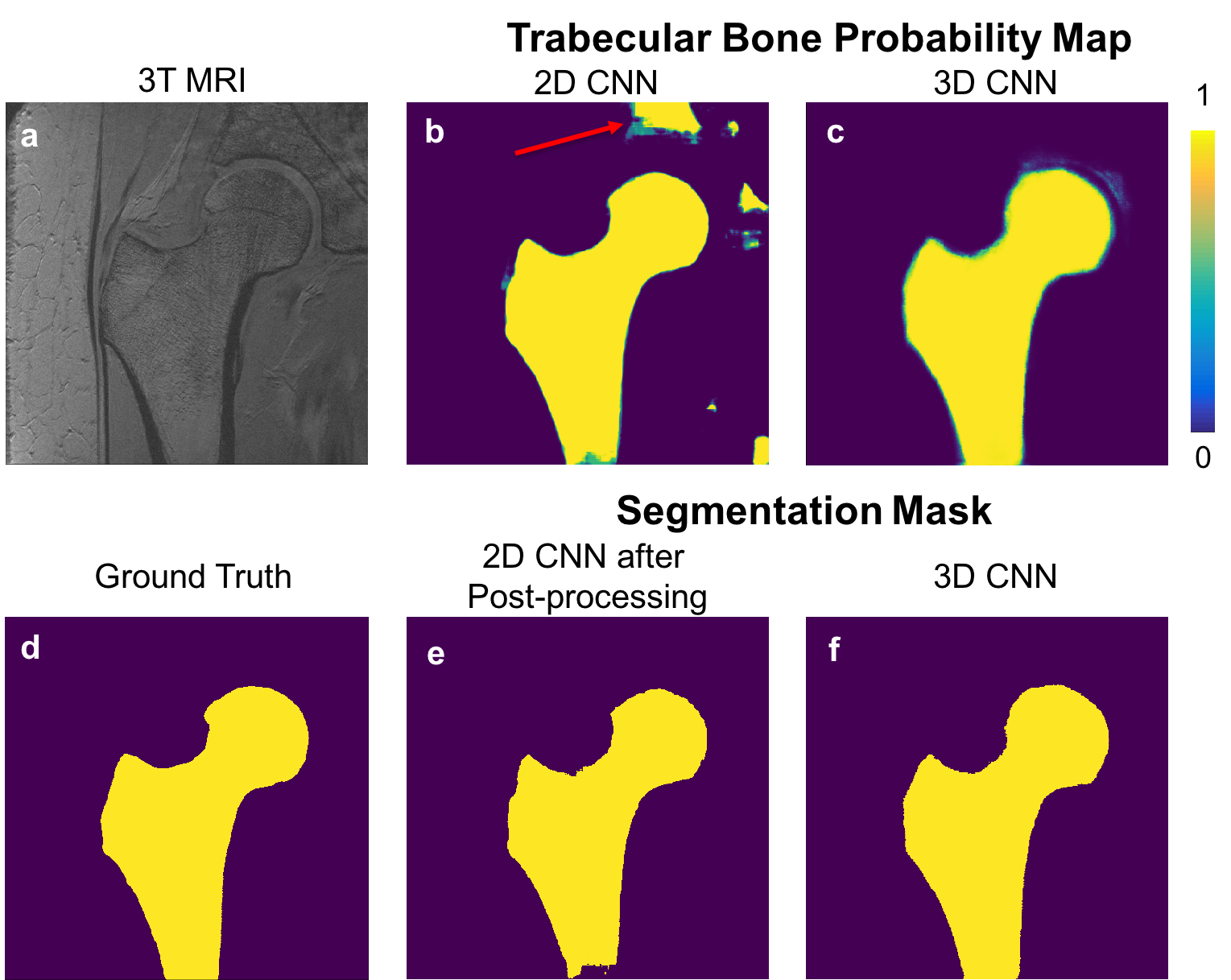}
\caption{An example of the results using 2D CNN and 3D CNN. 3T MRI of the proximal femur (a) is shown with the ground truth/hand segmentation mask (d). The probability map produced by 2D CNN is presented in (b) and corresponding segmentation mask after post-processing is presented in (e). Probability map produced by 3D CNN is presented in (c) and corresponding segmentation mask obtained by thresholding without post-processing is presented in (f). Red arrow in (b) indicates a location which was misclassified by the 2D CNN. Misclassified regions were removed by post-processing using proximal femur connectivity and size prior information (e).}
\label{figure:5}
\end{figure}

Segmentation results on one of the subjects is shown in Figure \ref{figure:5}. The proximal femur bone probability map from the 2D CNN includes misclassified regions which are not part of the proximal femur (as indicated by the red arrow). Removing the small clusters of misclassified bone regions with post-processing clearly improved the segmentation accuracy and results in a well-connected 3D proximal femur (Fig. \ref{figure:5}e). However, there are still misclassified locations remain, e.g. the bottom part of the proximal femur. In contrast to the 2D CNN, the 3D CNN automatically captures the global connectivity of the proximal femur during CNN training. This results in better delineation of the proximal femur on the trabecular bone probability map (Fig. \ref{figure:5}c) which provides a segmentation mask resembling the ground truth with higher accuracy. Because of this, as opposed to the 2D CNN, the 3D CNN doesn't require additional post-processing step.

\subsection*{Computational Efficiency}
Training each epoch takes approximately 5 minutes and 7 minutes for the 2D CNN and 3D CNN (for networks with 32 feature maps and 4 layers), respectively. The total time required for inference for the segmentation of data from one subject with central 48 coronal slices (covering the proximal femur) was approximately 18 seconds and 5 seconds for 2D CNN and 3D CNN (for networks with 32 feature maps and 4 layers), respectively. The increase in the inference time on the 2D CNNs was due to the use of multiple patches (9 patches per 2D slice) for calculating the segmentation mask on the full field of view. 

\section*{Discussion}
We present a deep CNN for automatic proximal femur segmentation from structural MR images. The automatic segmentation results indicate that the requirement of expert knowledge on location specifications and training/time for segmentation of the proximal femur may be avoided using CNNs. A Deep CNN for automatic segmentation can help bringing the use of proximal femur MRI measurements closer to clinical practice, given that manual segmentation of hip MR images can require approximately 1.5-2 hours of effort for high resolution volumetric datasets. 

CNN-based automatic segmentation of MR images has been performed in the brain \cite{Wachinger2017}, including for brain tumors \cite{Pereira2016}, microbleeds \cite{Dou2016}, and skull stripping for brain extraction \cite{Kleesiek2016}. CNN-based automatic segmentation has also been used for the pancreas \cite{Roth2015} and for knee \cite{Prasoon2013,Liu2018}. In recent years, automated segmentation of the proximal femur from MR images using a CNN begin to emerge in workshops \cite{Zeng2017} and conferences \cite{Hallyburton2017a}. Our results confirm previous results and extends them by adding a value in two aspects: (i) increased number of subjects, and (ii) analysis of architectures using 2D or 3D convolution in the concept of automated segmentation of the proximal femur from MR images. In the future, we expect the number of imaging applications of CNNs to rapidly increase, especially given that there are publicly available software libraries such as Tensorflow to create CNNs and that the algorithm can be executed on a commercially available desktop computers. 

In our implementation of the segmentation algorithms, we used 2D convolutional kernels in the first approach (2D CNN)  which could be one of the reasons for misclassified bone regions. Even though information from consecutive slices are incorporated in 2D CNN model training, global connectivity of the proximal femur may not be modeled properly using 2D convolutions alone. Although we used post-processing to prevent misclassified small regions in 2D CNNs, the approach using 3D convolutional kernels (3D CNN) resulted in a better segmentation masks by directly modeling the 3D connectivity of the proximal femur during training. Avoiding post-processing step in an automatic segmentation algorithm is crucial for segmentation tasks that aim to identify multiple regions. CNNs with 3D convolutional networks are computationally more demanding and can result in higher overfitting due to the increased number of weights to train. In our experiments, we used the validation error for an early stopping criteria to overcome successfully possible overfitting.  

In the 2D CNN, similar to the original U-net paper \cite{Ronneberger2015}, mirrored images were used during inference for calculating the probability of each voxel being part of the proximal femur. This resulted in inferencing on multiple patches covering the image and averaging the probability to calculate the output segmentation mask. Mirrored images can also be used during training, which removes the necessity of multiple calculations for averaging during inference. However, the increase in the input size of the network can result in an increased training time and a higher GPU memory requirement. On the other hand, using mirrored images for modeling will reduce the time required by inference and post-processing for 2D CNNs with unpadded convolution. In the 2D CNNs, padded convolutions instead of unpadded ones, as done in 3D CNN, can be used to obtain segmentation outputs that have the same size as the input images. This will remove the necessity of extracting multiple patches for calculating multiple segmentation probability maps and averaging them during inference. 

This study has limitations. First, the dataset consisted only of 86 subjects. In the future, with a larger dataset, we expect the performance of the CNNs to improve. Second, even though we implemented multiple CNNs with different number of  feature maps and layers, the automatic advanced hyperparameter optimization \cite{Snoek2012} for the CNN training parameters was not implemented in the current study. In the future, the optimization of learning rate and the number of initial feature maps will be performed. We expect the misclassified proximal femur bone regions in 2D CNN will be mitigated; and in both network architectures  this optimization will provide superior segmentation results. Third, image segmentation is a fast growing field with new architectures and approaches presented each year. We limited CNN architectures demonstrated in this work to cover current fundamental architectures \cite{Ronneberger2015, Cicek2016}, in which their variants have been used extensively for biomedical image segmentation. Comparing our results with the recent architectural developments \cite{Milletari2016,Lieman-Sifry2017,Kayalibay2017,Wachinger2017} and using different loss functions \cite{Milletari2016,Brosch2016} instead of weighted cross-entropy is beyond the scope of this work. 

In conclusion, we compared two major CNN architectures that are being increasingly used for biomedical image segmentation. Our experiments demonstrated the improved performance obtained using FCN and 3D convolutions for automatic segmentation of the proximal femur. The automatic segmentation using CNNs has the potential to help bringing the use of structural MRI measurements into the clinical practice.

\section*{Methods}
\subsection*{Convolutional Neural Networks}
\begin{figure}[ht]
\centering
\includegraphics[width=\linewidth]{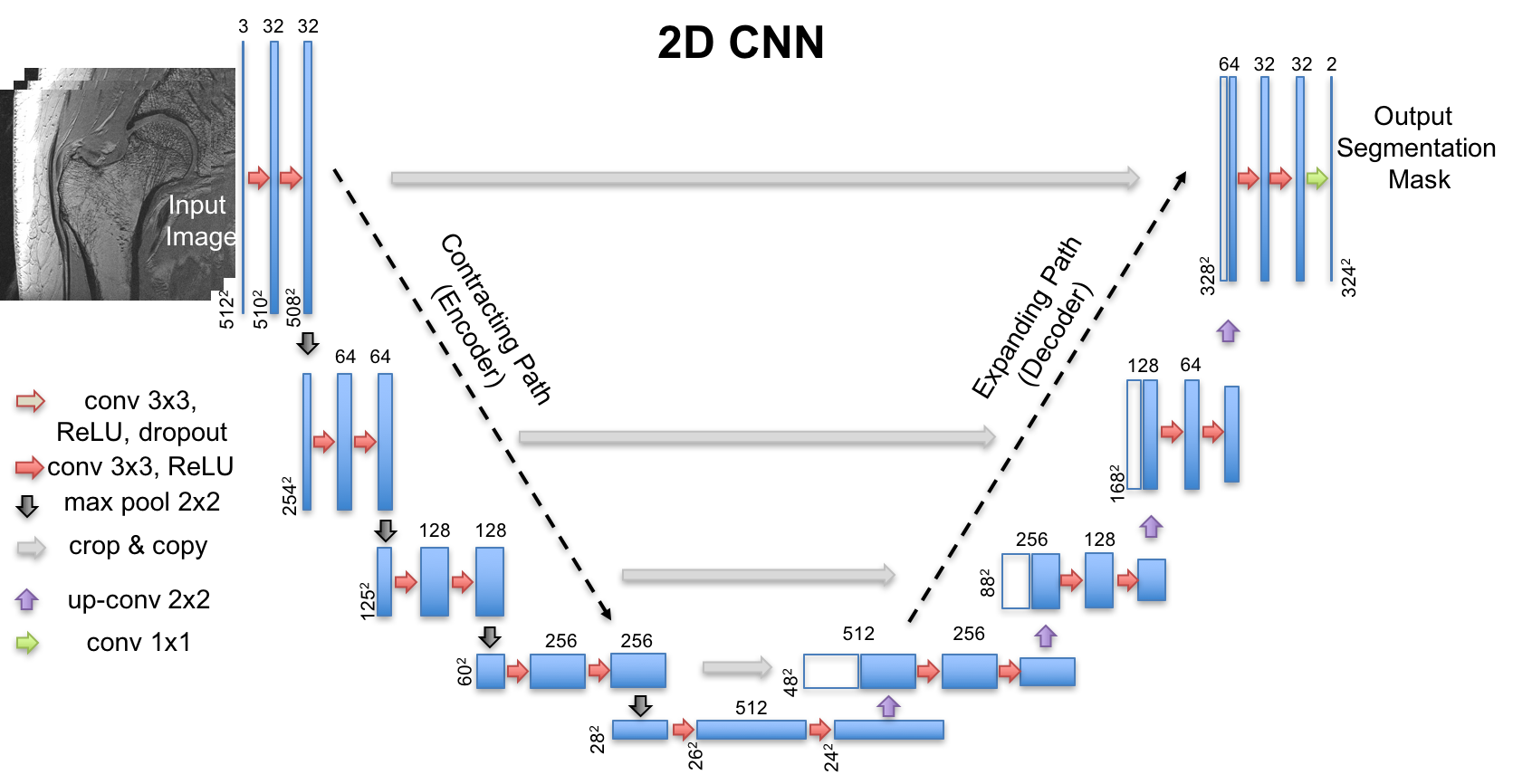}
\caption{CNN architecture of one of the 2D CNNs used in the paper. Blue rectangles represent feature maps with the size and the number of feature maps indicated. Different operations in the network are depicted by color-coded arrows. The architecture represented here contains 32 feature maps in the first and last layer of the network and 4 layers in the contracting/expanding paths.}
\label{figure:2}
\end{figure}

The first approach (2D CNN) uses a so-called U-net architecture \cite{Ronneberger2015} which was built upon a fully convolutional network (FCN) \cite{Long2014}. In the U-net architecture, the network uses a set of larger images as input and starts with a contracting path (encoder) similar to the conventional pyramidal CNN architectures \cite{Lecun1998}. Each pooling operation is followed by two convolutional layers with twice as many feature maps. After the contracting path, the network starts to expand in a way more or less symmetric to the contracting path (decoder), with some cropping and copying from the contracting path. This yields a U-shaped architecture (Fig. \ref{figure:2}). The output of the 2D CNN is a trabecular bone probability map of the center area of the input image. The size of the center area depends on the number of layers in the contracting/expanding paths. The second approach (3D CNN) is the extension of 2D CNN into three dimensions for volumetric segmentation using three-dimensional convolution, up-convolution and max-pooling layers \cite{Cicek2016}. In the 3D CNN, we use padded convolutions as opposed to unpadded ones proposed in \cite{Cicek2016} in order to provide a trabecular probability map of the whole image as an output. 

In all the CNNs, we use horizontal flipping for data augmentation \cite{Krizhevsky2012} since our dataset contained images from subjects who had been scanned either at the right hip or left hip. The initialization of the convolution kernel weights is known to be important to achieve convergence. In all experiments, we use the so-called Xavier \cite{Glorot2010} weight initialization method. The Xavier initializer is designed to keep the scale of the gradients roughly the same in all layers. This prevents the vanishing gradient \cite{Bengio1994} , enabling effective learning. As proposed in the original U-net article \cite{Ronneberger2015}, in the 2D CNN, we use unpadded 3x3 convolutions and 2x2 max-pooling operations with stride 2 to gradually decrease the size of the feature maps. In the expanding path, upsampling the feature map size is followed by an unpadded 2x2 up-convolution that halves the number of feature maps. For the 3D CNN, padded 3x3x3 convolutions and up-convolutions, 2x2x2 max-pooling with stride 2 are used in contrast to unpadded operations as proposed in \cite{Cicek2016} and \cite{Ronneberger2015}. Padded operations enable the size of the output trabecular bone mask to be equal to the input image size. This removes the requirement of using mirrored images during inference. For non-linearly transforming data within each layer of the CNN, rectifier linear unit (ReLU) \cite{Nair2010} is used as an activation function. ReLU is defined as $f(x)=\max(0,x)$. In the last layer of the CNN, we use softmax to compute the conditional distribution over the voxel label. 

The output of the softmax layer from the CNN is used to define a loss function which aims to minimize the error between the ground truth and the automatic segmentation via training. In our implementation, a loss function is defined as a negative log-probability of a target label (ground-truth) from an expert manually-segmented MR image. In medical images, the anatomical structure of interest usually occupies a small portion of the image. This potentially biases the CNN prediction towards background which constitutes the large portion of the images. To overcome this imbalanced class problem, we re-weighted the loss function during training. We achieve this by incorporating the number of proximal femur, $N_p$, and background, $N_b$, voxels into the loss value such that the error in voxels belonging to the trabecular bone are given more importance:
\begin{equation} \label{eq:1}
CE=-\frac{1}{N} \sum_{i=1}^N \Big(\frac{N_b}{N}y_i \log p_i + \frac{N_p}{N} (1-y_i)\log(1-p_i)\Big)
\end{equation}
where $N$ is the number of voxels, $y_i$ is a binary variable indicating if the trabecular bone is a correct prediction, $p_i$ is the probability of model prediction to be trabecular bone.

We use the Tensorflow \cite{Abadi2016} software library to implement CNNs. In the minimization of the loss function, we use adaptive moment estimation \cite{Kingma2014} (Adam). Parameters used in training the CNNs are outlined in Table \ref{table:1}. We perform experiments on a server using an NVIDIA 16GB Tesla P100 GPU card. For the 2D CNN, we used three consecutive slices and the segmentation mask from the center slice in order to capture some 3D connectivity information from 2D network architecture. 
 
\subsection*{Inference and Post-processing}
To predict the segmentation of the voxels in the border region of the images, we extrapolate the missing content by mirroring the input image during inference in experiments with the 2D CNNs. The probability of any voxel being trabecular bone can be calculated using multiple batches which covers that voxel at the center area of the patch. Because of this reason, during inference we use multiple patches for each voxel and average the probability of that voxel to calculate the probability of that voxel being trabecular bone. In total, we divide the mirrored image into 9 patches that cover the full mirrored image with an ordered overlap. For the 3D CNNs, mirroring of the images was not required due to the selection of padded convolutions in the network architecture.  

We perform basic post-processing on the segmentation results from the 2D CNNs to remove small clusters of misclassified bone regions. Since trabecular bone forms a 3D connected volume and covers the most number of voxels at the output of CNN, volumetric constraints are imposed by removing clusters with volumes smaller than the maximum volume of connected labels. The label corresponding to the maximum connected volume within 3D segmentation mask represents the proximal femur. This approach successfully removes those small clusters which were misclassified as proximal femur during the inference. Since using 3D convolution is capable of capturing 3D connectivity information of the trabecular bone accurately, this post-processing step is not required for the experiments based on the 3D CNNs. 

\subsection*{Dataset}
This study had institutional review board approval from New York University School of Medicine, and written informed consent was obtained from all subjects. The study was performed in accordance with all regulatory and ethical guidelines for the protection of human subjects by the National Institutes of Health. Images were obtained using commercial 3T MR scanner (Skyra, Siemens, Erlangen) with a 26-element radiofrequency coil setup (18-element Siemens commercial flexible array and 8-elements from the Siemens commercial spine array). High resolution proximal femur microarchitecture T1-weighted 3D fast low angle shot (3D FLASH) images were acquired with the following parameters: TR/TE= 31 / 4.92 ms; flip angle, 25$^{\circ}$; in-plane voxel size, 0.234 mm x 0.234 mm; section thickness, 1.5 mm; matrix size, 512x512; number of coronal sections, 60; acquisition time, 25 minutes 30 seconds; bandwidth, 200 Hz/pixel. High resolution acquisitions are required for resolving bone microarchitecture that is fundamental for accurate osteoporosis characterization. Using this imaging protocol, 86 post-menopausal women were scanned. Segmentation of the proximal femur was achieved by manual selection of the periosteal border of bone on MR images by an expert under the guidance of a musculoskeletal radiologist \cite{Link2012}. This resulted in two regions defined as trabecular bone of the proximal femur and the background. The central 48 coronal slices (covering 7.2 cm) were used for segmentation tasks covering the proximal femur and reducing the size of the input image especially for the 3D CNN.  Due to memory limitations of the GPU card, we resampled each slice of the MR images into 256x256 using bicubic spline interpolation, and used 16 and 32 initial feature maps for the 3D CNN. Analysis of the segmentation results were performed against the original (512x512) hand-segmented proximal femur masks.  

\subsection*{Model Selection}
Four-fold cross-validation is performed to assess the performance of different CNN architectures. Stratified random sampling is used to partition the sample into four disjoint groups. The first two groups have 21 subjects each, and the other two groups have 22 patients each. Each of the four groups serves as a validation set to assess the accuracy of a prediction model obtained from the other three groups combined as a training set. In this way, four separate segmentation models are derived, with each model is applied to segment the proximal femur in a validation set - data independent of the ones that is used to derive the model.

While training the CNNs, we use early stopping in order to prevent over-fitting and to enable fair comparison between different CNN architectures. Training is stopped when the accuracy on the validation set does not improve by $10^{-4}$ within the last 10 epochs. First 30 epochs are trained without early stopping. 

\subsection*{Evaluation}
Manual segmentations of the proximal femur  were used as the ground truth to evaluate different CNN structures. We define voxels within the proximal femur and background voxels as positive and negative outcomes, respectively.The performance of CNNs are evaluated using ROC and PRC analysis, DSC, sensitivity/recall, and precision. The DSC metric \cite{Dice1945}, also known as F1-score, measures the similarity/overlap between manual and automatic segmentations. DSC metric is the most widely used metric when validating medical volume segmentations \cite{Taha2015}, and it is defined as:
\begin{equation} \label{eq:2}
DSC=2TP/(FP+2TP+FN)
\end{equation}
where TP, FP, and FN are detected number of true positives, false positives and false negatives, respectively. Sensitivity/recall measures the portion of proximal femur bone voxels in the ground truth that are also identified as a proximal femur bone voxel by the automatic segmentation. Sensitivity/recall is defined as:
\begin{equation} \label{eq:3}
sensitivity/recall=TP/(TP+FN)
\end{equation}
Similarly, specificity measures the portion of background voxels in the ground truth that are also identified as a background voxel by the automatic segmentation. Specificity is defined as:
\begin{equation} \label{eq:4}
specificity=TN/(TN+FP)
\end{equation}
Lastly, precision, also known as positive predictive value (PPV), measures the proportion of trabecular bone voxels in the ground truth and voxels identified as trabecular bone by the automatic segmentation. It is defined as:
\begin{equation} \label{eq:5}
precision(PPV)=TP/(TP+FP)
\end{equation}
ROC curve analysis provides a means of evaluating the performance of automatic segmentation algorithms and selecting a suitable decision threshold. We use the area under the PRC (AP) as a measure of classifier’s performance for comparing different CNNs. The output of a CNN defines the probability of a voxel belonging within the proximal femur. Using PRC analysis, the optimal threshold is selected for each CNN to distinguish proximal femur bone voxels from background when comparing the performance of CNNs. The optimal operating point for each CNN was selected by choosing the point on the PRC that has the smallest Euclidean distance to the maximum precision and recall. The voxels having higher probabilities then selected threshold is predicted as belonging within the proximal femur and the rest as background. 

\subsection*{Data Availability}
The datasets generated during and/or analyzed during the current study are available from the corresponding author on reasonable request.

\bibliography{sample}

\section*{Acknowledgements}
This work was supported in part by NIH R01 AR066008 and NIH R01 AR070131 and was performed under the rubric of the Center for Advanced Imaging Innovation and Research (CAI2R, www.cai2r.net), an NIBIB Biomedical Technology Resource Center (NIH P41 EB017183). We gratefully acknowledge the support of NVIDIA Corporation with the donation of a GPU for this research.

\section*{Author contributions statement}
C.M.D.: study concept and design, experiments, analysis of the results, manuscript preparation
S.X.: literature research, 3D CNN implementation, data analysis
S.H: literature research, data preparation  
A.W.: data acquisition and segmentation
S.H.: study concept and patient recruitment
K.C.: study design and manuscript editing
G.C: data acquisition, data segmentation and manuscript editing

\section*{Additional information}
\textbf{Competing financial interests:} G.C. has a pending patent application. The other authors do not have conflict of interests to disclose. 

\begin{table}[ht]
\centering
\begin{tabular}{|l|l|l|}
\hline \hline
Phase & Parameter & Value\\
\hline
Initialization & Weights & Xavier\\
 & Bias & .10\\
Training & Input Image Size - 2D CNN & 512x512x3\\
& Input Image Size - 3D CNN & 512x512x48 \\
& Optimizer & Adam\\
& Batch Size & 1\\
& Learning Rate & 5e-5\\
\hline \hline
\end{tabular}
\caption{\label{table:1}Hyperparameters used for CNN training.}
\end{table}

\begin{table}[ht]
\centering
\begin{tabular}{c|c|c|c}
\hline \hline
Network & DSC & Precision & Recall\\
\hline
2D CNN, F:64, L:4 	&	0.864$\pm$0.044	&	0.872$\pm$0.061	&	0.860$\pm$0.060 \\
2D CNN, F:64, L:3 	&	0.886$\pm$0.055	&	0.890$\pm$0.080	&	0.889$\pm$0.056 \\
2D CNN PP, F:64, L:4	& 	0.920$\pm$0.040	&	0.991$\pm$0.010	&	0.861$\pm$0.060 \\
2D CNN PP, F:64, L:3	& 	0.935$\pm$0.034	&	0.990$\pm$0.010	&	0.889$\pm$0.056 \\
3D CNN, F:32, L:4     & 	0.940$\pm$0.054	&	0.946$\pm$0.024	&	0.939$\pm$0.082 \\
3D CNN, F:32, L:3     & 	0.939$\pm$0.041	&	0.943$\pm$0.023	&	0.938$\pm$0.072 \\
3D CNN, F:16, L:4     & 	0.930$\pm$0.057	&	0.937$\pm$0.027	&	0.930$\pm$0.085 \\
3D CNN, F:16, L:3     & 	0.924$\pm$0.048	&	0.929$\pm$0.029	&	0.924$\pm$0.082 \\
\hline \hline
\end{tabular}
\caption{\label{table:2}Cross-validation results of different network architectures for the segmentation of proximal femur. Segmentation algorithms including post-processing represented by PP. F is the number of initial feature maps, L is the number of layers.}
\end{table}

\end{document}